\documentclass{article}

\usepackage{PRIMEarxiv}

\usepackage[utf8]{inputenc} 
\usepackage[T1]{fontenc}    
\usepackage{hyperref}       
\usepackage{url}            
\usepackage{booktabs}       
\usepackage{amsfonts}       
\usepackage{nicefrac}       
\usepackage{microtype}      
\usepackage{lipsum}
\usepackage{fancyhdr}       
\usepackage{graphicx}       
\graphicspath{{media/}}     

\title{Towards Zero-Shot and Few-Shot Table Question Answering
Using GPT-3
\thanks{\textit{This work was done when Pragya Srivastava was an Intern at Microsoft Research Labs, India}}
}

\author{
  Pragya Srivastava \\
  Indian Institute of Technology, Delhi\\
   New Delhi, India\\
  \texttt{\{ph1190646\}@iitd.ac.in} \\
   \And
  Tanuja Ganu, Saikat Guha \\
  Microsoft Research Labs \\
  Bengaluru, India\\
  \texttt{\{taganu, saikat\}@microsoft.com} \\
}

\begin{document}
\maketitle

\begin{abstract}
We present very early results on using GPT-3 to perform question answering on tabular data. We find that stock pre-trained GPT-3 is able to zero-shot learn the table structure from a serialized JSON array-of-arrays representation, and able to answer lookup queries and simple comparison questions in natural language without any fine-tuning. We further find that simple prompt engineering to include few-shot static Q\&A examples significantly improves accuracy. Lastly, we find that intermixing passage text improves accuracy even further on heterogeneous data. We apply our approach on a novel dataset of simple tables in newspaper infographics with promising results. Overall, we find much cause for optimism in this basic approach.
\end{abstract}


\section{Introduction}

Foundational models like GPT-3~\cite{NEURIPS2020_1457c0d6} have revolutionized natural language understanding. Researchers discovered that GPT-3 had unexpectedly gained the ability to understand and synthesize source code, albeit with low accuracy. This observation resulted in a concerted effort to fine-tune GPT-3 on large codebases, resulting in Codex~\cite{https://doi.org/10.48550/arxiv.2107.03374}. We ask a similar question of whether GPT-3 has unexpectedly gained the ability to understand tabular data.

\begin{figure*}[t]
\centering
\includegraphics[height=3in]{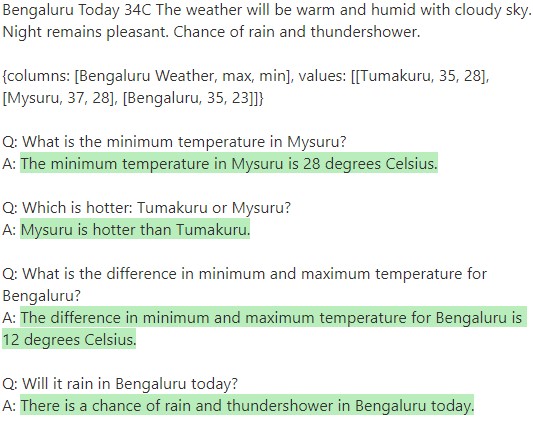}
\caption{GPT-3 prompt containing tabular data and questions in black-on-white text, followed by GPT-3 responses in black-on-green.}
\label{fig:interaction}
\end{figure*}

We prompted unmodified pre-trained GPT-3 with tabular data and simple questions and discovered, quite unexpectedly, that GPT-3 has indeed developed the capability to understand tabular data and answer simple questions. Consider for instance the interaction in Figure~\ref{fig:interaction} where the black-on-white text is the prompt to GPT-3 and the black-on-green text is the GPT-3 response. Notice the serialized JSON representation of the table, intermixed with unstructured passage text. GPT-3 was able to successfully locate the table, understand its structure, index into the correct cells or passage text to answer the questions posed, and formulate a response in natural language, all without any fine-tuning. As we report later, this basic zero-shot approach has some limitations. We then engineered our prompts using few-shot static Q\&A examples and found significant improvement in accuracy. Overall, we anticipate our findings will spark new approaches to table representation learning and question answering using GPT-3.

This paper reports on our methodology and early results with using GPT-3 on tabular data. Section~\ref{sec:related} sets background and related work. Section~\ref{sec:methodology} presents our methodology. Section~\ref{sec:experiments} reports on our experiment results. Section~\ref{sec:discussion} discusses ongoing work. And Section~\ref{sec:conclusion} concludes this paper.


 

\section{Background and Related Work}
\label{sec:related}

A large body of work has explored novel model architectures for table representation learning and question answering~\cite{herzig-etal-2020-tapas,https://doi.org/10.48550/arxiv.2005.08314,liu2021tapex,chen-etal-2020-hybridqa,zhongSeq2SQL2017}. Two recent examples include TAPAS and HybridQA. TAPAS~\cite{herzig-etal-2020-tapas} trains a BERT-based model with additional position embeddings to encode tabular structure. HybridQA~\cite{chen-etal-2020-hybridqa} trains HyBrider, a model that combines table data and related passage data. In contrast, our work applies the unmodified pre-trained GPT-3 model to specially engineered prompts. 


\begin{figure*}[h]
\centering
\includegraphics[width=\textwidth]{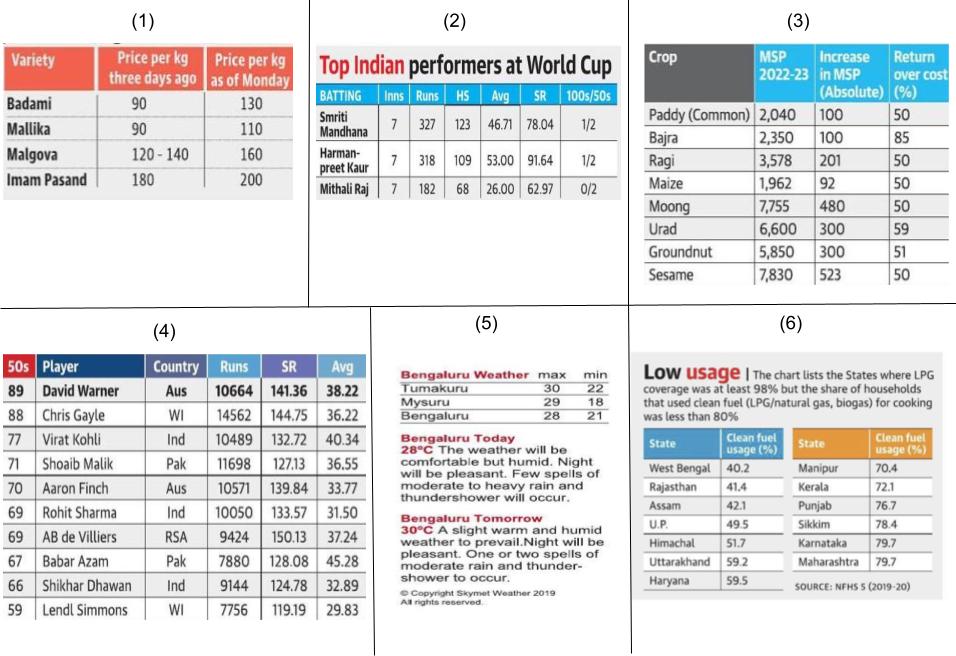}
\caption{A sampling of infographic tables from our dataset.}
\label{fig:fig_tables}
\end{figure*}

\begin{table*}[h]
\centering
\footnotesize
\label{table:table_3}
\begin{tabular}{ |p{0.5cm}|p{4cm}|p{1.5cm}|p{3.5cm}|p{2.5cm}| }
\hline
\textbf{TID} &\textbf{Question} & \textbf{Expected} & \textbf{GPT-3 (Zero-Shot)} & \textbf{GPT-3 (Few-Shot)} \\
\hline
1 & What is the lowest value in the column "Price per kg as of Monday"? & 110 & The lowest value in the "Price per kg as of Monday" column is 90. & 110 \\
\hline
1 & Which variety of Mango saw the least hike in price and what is the hike? & Mallika,Imam Pasand, 20 & The variety of mango that saw the least price hike is Badami, with a price hike of 40\% & Mallika, 20 \\
\hline
2 & What is the highest strike rate of any player? & 91.64 & 78.04 & 91.64 \\
\hline
2 & Sort the players in increasing order of their average scores. & Mithali Raj, Smriti Mandhana, Harmanpreet Kaur & [[Mithali Raj 7, 7, 182, 68, 26.0, 62.97, 0/2], [Harmanpreet Kaur , 7, 318, 109, 53.0, 91.64, 1/2], [Smriti Mandhana , 7, 327, 123, 46.71, 78.04, 1/2]] & Mithali Raj, Smriti Mandhana, Harmanpreet Kaur \\
\hline
3 & Which crop has the highest return over cost percentage? & Bajra & Sesame has the highest return over cost percentage at 50\% & Bajra \\

\hline
3 & What was the MSP of Bajra in "2021-22"? & 2250 &  The MSP of Bajra in 2021-22 was 2350 & 2,250 \\
\hline

4 & How many Indian players are above Virat Kohli in terms of strike rate? & 1 & 3 & 1 \\
\hline
4 & Which country's player has the highest strike rate? & RSA & The player with the highest strike rate is Aaron Finch of Australia & RSA \\
\hline
5 & Which city has the least difference between minimum and maximum temperatures? & Bengaluru & Mysuru & Bengaluru \\
\hline
5 & Which city has the least maximum temperature? & Bengaluru & Mysuru & Bengaluru \\
\hline
6 & What is the coverage of LPG in Maharashtra? & 98\% & 79.7\% & 98\%  \\
\hline
\end{tabular}
\caption{Failure cases for zero-shot Q\&A, which is addressed using few-shot learning. TID refers to table ID in Figure~\ref{fig:fig_tables}}
\end{table*}

\section{Methodology}
\label{sec:methodology}

In this section we describe our dataset generation process, zero-shot methodology, and few-shot methodology.

\textbf{Dataset Generation.} We use Bing image search to find images of digital newspaper pages. We then use a fine-tuned Faster RCNN~\cite{NIPS2015_14bfa6bb} object detection model, pre-trained on the COCO dataset~\cite{Lin2014MicrosoftCC}, to segment out infographics. We then use Azure Form Recognizer to recognize tables in these infographics and nearby passage text. Finally, we classify the tables into weather, sports, financial and others using RoBERTa~\cite{Liu2019RoBERTaAR} fine-tuned on our dataset.

\textbf{Zero-Shot Learning.} We construct GPT-3 prompts in the format \texttt{<passage text> <table data> Q: <question> A:} as shown in Figure~\ref{fig:interaction}. Passage text is unprocessed OCR output of nearby text. Table data is the unprocessed JSON output from Azure Form Recognizer. Question is the natural language question we seek to answer generated as explained next.

In order to generate the questions and (ground-truth) answers, we use the pretrained Google T5 model~\cite{Raffel2020ExploringTL} fine-tuned on SQuAD~\cite{rajpurkar-etal-2016-squad} data split from \cite{zhou2017neural}. See Appendix~\ref{sec:qaappendix} for details.  

\textbf{Few-Shot Learning.} We generate synthetic tables and Q \& A examples using the approach in~\cite{pan-etal-2021-unsupervised}. This synthetic example is encoded as \texttt{<synthetic data> Q: <synthetic question> A: <synthetic answer> ...} and prepended to the prompt specified above. Note that the synthetic example is in the same category (i.e. weather, sports) as the target table, but the synthetic data is otherwise unrelated to the target table.

We generate synthetic question and answers using the Google T5 model. We further augment our synthetic Q\&A examples by including simple statistical questions like minimum, maximum, or mean values from a column (the end-to-end process is described in appendix~\ref{sec:qaappendix}).

\section{Experiments}
\label{sec:experiments}

We collected a small dataset of 12 infographic tables using the methodology described earlier as our test set; see Figure~\ref{fig:fig_tables} for samples. We also created a set of 200 synthetic Q\&A pairs for the few-shot learning experiments using 37 inforgraphic tables, as described earlier. This sections reports on our experimental results as compared to TAPAS~\cite{herzig-etal-2020-tapas} fine-tuned on our dataset.

\begin{table}
  \centering
  \begin{tabular}{ll}
    \cmidrule(r){1-2}
    Model     & Accuracy \\
    \midrule
     TAPAS  & $   18.4$ \\
    GPT-3 (Zero-Shot) & $     73.6$\\
    GPT-3 (Few-Shot)  & $     86.8$\\ 
    \bottomrule
  \end{tabular}
 \caption{Comparison of fine-tuned TAPAS and GPT-3 zero-shot and few-shot on our small dataset}
  \label{table:results}
\end{table}

Table~\ref{table:results} compares GPT-3 performance to TAPAS. We find that zero-shot learning achieves an accuracy of 73.6\% compared to fine-tuned TAPAS accuracy of 18.4\% on our small dataset. Table~\ref{fig:fig_tables} samples some Q\&A that GPT-3 with zero-shot learning answers incorrectly. GPT-3 with few-shot learning correctly answers these examples and achieves an accuracy of 86.8\%. That being said, we caution the reader due to the small size of our dataset.

\section{Discussion}
\label{sec:discussion}

This paper presents very early results, albeit results that hold much promise. We are currently focusing on the following aspects.

\textbf{Benchmarks.} We are working on applying our approach to benchmark datasets including Spider, WikiTableQuestions, and WikiSQL ~\cite{https://doi.org/10.48550/arxiv.1809.08887,https://doi.org/10.48550/arxiv.1508.00305,zhongSeq2SQL2017}. We are also working on comparing against additional approaches including TaBERT, TAPEX and HybridQA~\cite{https://doi.org/10.48550/arxiv.2005.08314,liu2021tapex,chen-etal-2020-hybridqa}. 

\textbf{Scale.} We are looking to scale the novel newspaper infographic dataset and release it publicly. We believe it is a unique dataset, geared towards the general non-technical audience, fine-tuning on which might unlock many applications.

\textbf{Applications.} We are looking to serve the visually impaired community for whom infographics in newspapers and magazines are inaccessible. We are also looking at applications in K-6 education helping young children be able to query and understand data and develop critical data interrogation skills.

\section{Conclusion}
\label{sec:conclusion}

We present first steps towards using foundational models for learning table representations and performing question-answering on tabular data. We found that GPT-3's zero-shot performance is already quite good, and can be significantly improved upon with synthetic few-shot examples. We presented a preliminary evaluation of our approach on a novel dataset comprising simple tables from newspaper infographics. Overall, we conclude that using foundational models is a promising approach for table representation learning and question-answering.


 

\bibliographystyle{unsrt}  
\bibliography{main} 

\begin{thebibliography}{10}

\bibitem{NEURIPS2020_1457c0d6}
Tom Brown, Benjamin Mann, Nick Ryder, Melanie Subbiah, Jared~D Kaplan, Prafulla
  Dhariwal, Arvind Neelakantan, Pranav Shyam, Girish Sastry, Amanda Askell,
  Sandhini Agarwal, Ariel Herbert-Voss, Gretchen Krueger, Tom Henighan, Rewon
  Child, Aditya Ramesh, Daniel Ziegler, Jeffrey Wu, Clemens Winter, Chris
  Hesse, Mark Chen, Eric Sigler, Mateusz Litwin, Scott Gray, Benjamin Chess,
  Jack Clark, Christopher Berner, Sam McCandlish, Alec Radford, Ilya Sutskever,
  and Dario Amodei.
\newblock Language models are few-shot learners.
\newblock In H.~Larochelle, M.~Ranzato, R.~Hadsell, M.F. Balcan, and H.~Lin,
  editors, {\em Advances in Neural Information Processing Systems}, volume~33,
  pages 1877--1901. Curran Associates, Inc., 2020.

\bibitem{https://doi.org/10.48550/arxiv.2107.03374}
Mark Chen, Jerry Tworek, Heewoo Jun, Qiming Yuan, Henrique Ponde de~Oliveira
  Pinto, Jared Kaplan, Harri Edwards, Yuri Burda, Nicholas Joseph, Greg
  Brockman, Alex Ray, Raul Puri, Gretchen Krueger, Michael Petrov, Heidy
  Khlaaf, Girish Sastry, Pamela Mishkin, Brooke Chan, Scott Gray, Nick Ryder,
  Mikhail Pavlov, Alethea Power, Lukasz Kaiser, Mohammad Bavarian, Clemens
  Winter, Philippe Tillet, Felipe~Petroski Such, Dave Cummings, Matthias
  Plappert, Fotios Chantzis, Elizabeth Barnes, Ariel Herbert-Voss,
  William~Hebgen Guss, Alex Nichol, Alex Paino, Nikolas Tezak, Jie Tang, Igor
  Babuschkin, Suchir Balaji, Shantanu Jain, William Saunders, Christopher
  Hesse, Andrew~N. Carr, Jan Leike, Josh Achiam, Vedant Misra, Evan Morikawa,
  Alec Radford, Matthew Knight, Miles Brundage, Mira Murati, Katie Mayer, Peter
  Welinder, Bob McGrew, Dario Amodei, Sam McCandlish, Ilya Sutskever, and
  Wojciech Zaremba.
\newblock Evaluating large language models trained on code, 2021.

\bibitem{herzig-etal-2020-tapas}
Jonathan Herzig, Pawel~Krzysztof Nowak, Thomas M{\"u}ller, Francesco Piccinno,
  and Julian Eisenschlos.
\newblock {T}a{P}as: Weakly supervised table parsing via pre-training.
\newblock In {\em Proceedings of the 58th Annual Meeting of the Association for
  Computational Linguistics}, pages 4320--4333, Online, July 2020. Association
  for Computational Linguistics.

\bibitem{https://doi.org/10.48550/arxiv.2005.08314}
Pengcheng Yin, Graham Neubig, Wen-tau Yih, and Sebastian Riedel.
\newblock Tabert: Pretraining for joint understanding of textual and tabular
  data, 2020.

\bibitem{liu2021tapex}
Qian Liu, Bei Chen, Jiaqi Guo, Zeqi Lin, and Jian guang Lou.
\newblock Tapex: Table pre-training via learning a neural sql executor, 2021.

\bibitem{chen-etal-2020-hybridqa}
Wenhu Chen, Hanwen Zha, Zhiyu Chen, Wenhan Xiong, Hong Wang, and William~Yang
  Wang.
\newblock {H}ybrid{QA}: A dataset of multi-hop question answering over tabular
  and textual data.
\newblock In {\em Findings of the Association for Computational Linguistics:
  EMNLP 2020}, pages 1026--1036, Online, November 2020. Association for
  Computational Linguistics.

\bibitem{zhongSeq2SQL2017}
Victor Zhong, Caiming Xiong, and Richard Socher.
\newblock Seq2sql: Generating structured queries from natural language using
  reinforcement learning.
\newblock {\em CoRR}, abs/1709.00103, 2017.

\bibitem{NIPS2015_14bfa6bb}
Shaoqing Ren, Kaiming He, Ross Girshick, and Jian Sun.
\newblock Faster r-cnn: Towards real-time object detection with region proposal
  networks.
\newblock In C.~Cortes, N.~Lawrence, D.~Lee, M.~Sugiyama, and R.~Garnett,
  editors, {\em Advances in Neural Information Processing Systems}, volume~28.
  Curran Associates, Inc., 2015.

\bibitem{Lin2014MicrosoftCC}
Tsung-Yi Lin, Michael Maire, Serge~J. Belongie, James Hays, Pietro Perona, Deva
  Ramanan, Piotr Doll{\'a}r, and C.~Lawrence Zitnick.
\newblock Microsoft coco: Common objects in context.
\newblock In {\em ECCV}, 2014.

\bibitem{Liu2019RoBERTaAR}
Yinhan Liu, Myle Ott, Naman Goyal, Jingfei Du, Mandar Joshi, Danqi Chen, Omer
  Levy, Mike Lewis, Luke Zettlemoyer, and Veselin Stoyanov.
\newblock Roberta: A robustly optimized bert pretraining approach.
\newblock {\em ArXiv}, abs/1907.11692, 2019.

\bibitem{Raffel2020ExploringTL}
Colin Raffel, Noam~M. Shazeer, Adam Roberts, Katherine Lee, Sharan Narang,
  Michael Matena, Yanqi Zhou, Wei Li, and Peter~J. Liu.
\newblock Exploring the limits of transfer learning with a unified text-to-text
  transformer.
\newblock {\em ArXiv}, abs/1910.10683, 2020.

\bibitem{rajpurkar-etal-2016-squad}
Pranav Rajpurkar, Jian Zhang, Konstantin Lopyrev, and Percy Liang.
\newblock {SQ}u{AD}: 100,000+ questions for machine comprehension of text.
\newblock In {\em Proceedings of the 2016 Conference on Empirical Methods in
  Natural Language Processing}, pages 2383--2392, Austin, Texas, November 2016.
  Association for Computational Linguistics.

\bibitem{zhou2017neural}
Qingyu Zhou, Nan Yang, Furu Wei, Chuanqi Tan, Hangbo Bao, and Ming Zhou.
\newblock Neural question generation from text: A preliminary study.
\newblock {\em arXiv preprint arXiv:1704.01792}, 2017.

\bibitem{pan-etal-2021-unsupervised}
Liangming Pan, Wenhu Chen, Wenhan Xiong, Min-Yen Kan, and William~Yang Wang.
\newblock Unsupervised multi-hop question answering by question generation.
\newblock In {\em Proceedings of the 2021 Conference of the North American
  Chapter of the Association for Computational Linguistics: Human Language
  Technologies}, pages 5866--5880, Online, June 2021. Association for
  Computational Linguistics.

\bibitem{https://doi.org/10.48550/arxiv.1809.08887}
Tao Yu, Rui Zhang, Kai Yang, Michihiro Yasunaga, Dongxu Wang, Zifan Li, James
  Ma, Irene Li, Qingning Yao, Shanelle Roman, Zilin Zhang, and Dragomir Radev.
\newblock Spider: A large-scale human-labeled dataset for complex and
  cross-domain semantic parsing and text-to-sql task, 2018.

\bibitem{https://doi.org/10.48550/arxiv.1508.00305}
Panupong Pasupat and Percy Liang.
\newblock Compositional semantic parsing on semi-structured tables, 2015.

\end{thebibliography}

\newpage

\appendix

\section{Generating Questions and Answers}
\label{sec:qaappendix}

\begin{figure}
    \centering
    \includegraphics[width=15cm]{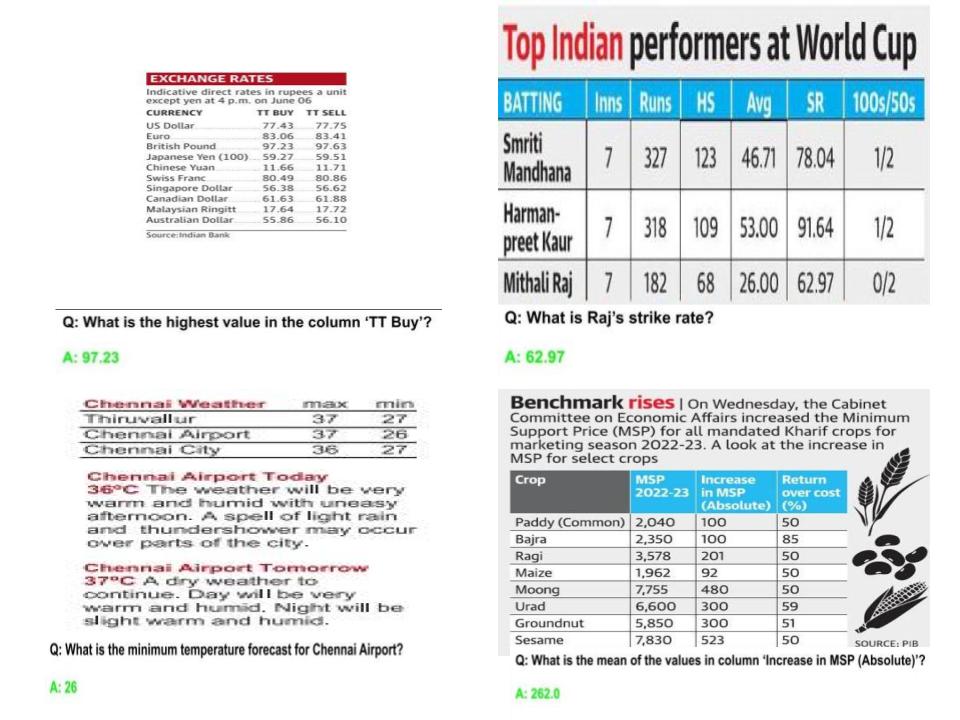}
    \caption{Examples of the generated QA pairs which serve as in-context examples to GPT-3}
    \label{fig:examples}
\end{figure}

Figure ~\ref{fig:examples} shows some generated questions and answers. We use these generated questions and answers in two places as explained in Section~\ref{sec:methodology}. First, we use them as the ground-truth in our primary dataset. And second, we use them as synthetic examples when performing our few-shot approach. We generate these questions and answers as follows.

\begin{figure}[b]
\centering
\includegraphics[height=3in]{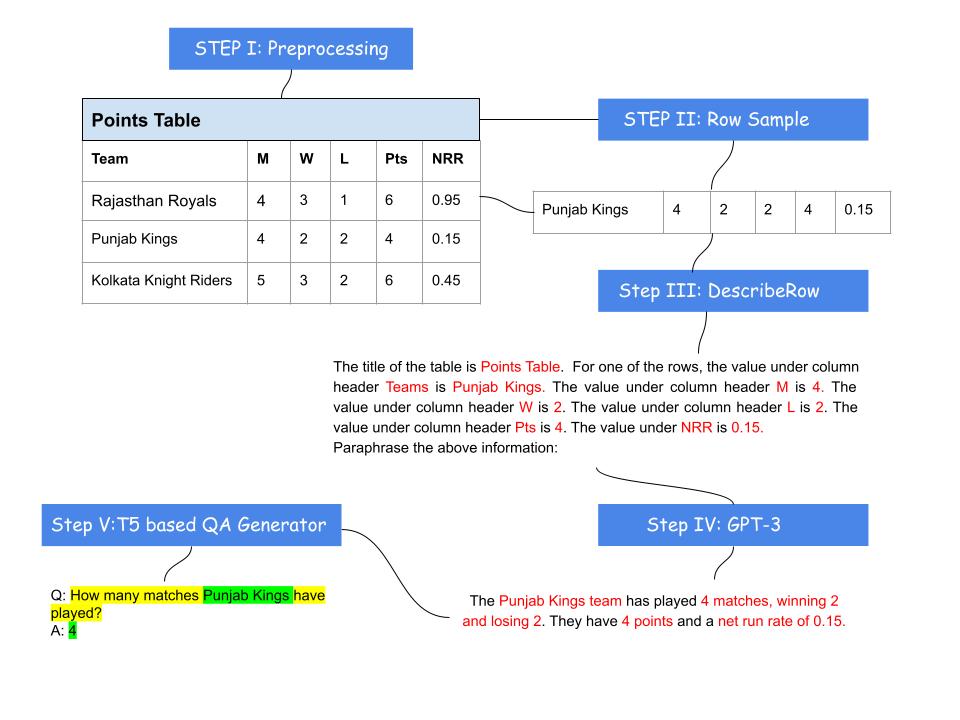}
\caption{An illustration of the end-to-end framework for Question-Answer pair generation}
\label{fig:qageneration}
\end{figure}

For each table, we randomly sample a row from it and apply a defined \textit{DescribeRow} operation where we apply a fixed template to describe only the sampled row in natural text. The fixed template is shown in black text in Figure~\ref{fig:qageneration} and the placeholders replaced with data from the row is shown in red. We then use GPT-3 to summarize the templated row description. Note this use of GPT-3 during dataset generation is unrelated to our primary use GPT-3 for question-answering. 

Using the row summary descriptions generated in the previous step, we generate question-answering pairs over tables and their linked texts using the same methodology as \cite{pan-etal-2021-unsupervised}. We use the pretrained Google T5 model~\cite{Raffel2020ExploringTL} fine-tuned on SQuAD~\cite{rajpurkar-etal-2016-squad} data split from \cite{zhou2017neural} to finetune Google's T5 model. Given the SQuAD training set of context-question-answer triples D = {(\textbf{c},\textbf{q},\textbf{a})}, we jointly fine-tune the model on the task of 1) \emph{QGwithAns} aims to generate a question q with a as the answer, given (\textbf{c},\textbf{a}) as inputs. 2) \emph{QGwithEnt} aims to generate a question q that contains a specific entity e, given (c,e) as inputs. We implement this based on the pretrained T5 model\footnote{\url{https://github.com/patil-suraj/question\_generation}} and execute the end-to-end question generation process (answer agnostic).

\begin{figure}
\centering
\includegraphics[height=1in]{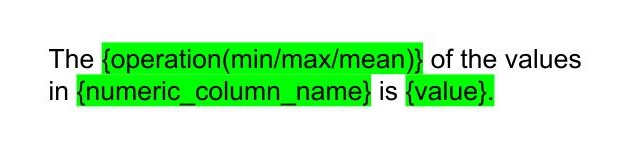}
\caption{Template for simple statistical Q\&A}
\label{fig:fig_temp}
\end{figure}

We further augment our generated Q\&A to include simple table statistics for numeric columns generated using the template in Figure~\ref{fig:fig_temp}. The black-on-green placeholders are replaced with the statistical operation (min, max, or mean) and the name of the numeric column and static value.

%

\end{document}